\documentclass{article}

\usepackage[preprint]{neurips_2023}

\usepackage[utf8]{inputenc} % allow utf-8 input
\usepackage[T1]{fontenc}    % use 8-bit T1 fonts
\usepackage{hyperref}       % hyperlinks
\usepackage{url}            % simple URL typesetting
\usepackage{booktabs}       % professional-quality tables
\usepackage{amsfonts}       % blackboard math symbols
\usepackage{nicefrac}       % compact symbols for 1/2, etc.
\usepackage{microtype}      % microtypography
\usepackage{xcolor}         % colors

\usepackage{graphicx}
\usepackage{caption}
\usepackage{subcaption}
\usepackage{amsmath}
\usepackage{cleveref}

\title{Verbosity Bias in Preference Labeling \\ by Large Language Models}

\author{%
  Keita Saito\thanks{Research done during internship at LY Corp.} \\
  University of Tsukuba \& RIKEN AIP\\
  Tsukuba, Ibaraki 305-8573, Japan \\
  \texttt{keita.saito@bbo.cs.tsukuba.ac.jp} \\
  \And
  Akifumi Wachi \\
  LY Corporation \\
  Chiyoda-ku, Tokyo 102-8282, Japan \\
  \texttt{akifumi.wachi@lycorp.co.jp} \\
  \AND
  Koki Wataoka \\
  LY Corporation \\
  Chiyoda-ku, Tokyo 102-8282, Japan \\
  \texttt{koki.wataoka@lycorp.co.jp} \\
  \And
  Youhei Akimoto \\
  University of Tsukuba \& RIKEN AIP\\
  Tsukuba, Ibaraki 305-8573, Japan \\
  \texttt{akimoto@cs.tsukuba.ac.jp}
}

\begin{document}

\maketitle

\begin{abstract}
In recent years, Large Language Models (LLMs) have witnessed a remarkable surge in prevalence, altering the landscape of natural language processing and machine learning. One key factor in improving the performance of LLMs is alignment with humans achieved with Reinforcement Learning from Human Feedback (RLHF), as for many LLMs such as GPT-4, Bard, etc. In addition, recent studies are investigating the replacement of human feedback with feedback from other LLMs named Reinforcement Learning from AI Feedback (RLAIF). We examine the biases that come along with evaluating LLMs with other LLMs and take a closer look into verbosity bias – a bias where LLMs sometimes prefer more verbose answers even if they have similar qualities. We see that in our problem setting, GPT-4 prefers longer answers more than humans. We also propose a metric to measure this bias.
\end{abstract}

\section{Introduction}

% LLMs have become widespread (chatbots, summary generation, etc.)
Large Language Models (LLMs) have made tremendous strides in recent years and continue to gain popularity~\citep{zhao2023survey}. With its growing size in network parameters, its wide application ranges from conventional natural language processing tasks such as chat-bots, summarization, and translation, to other applications beyond its original intended use such as search engines, programming assistance, and foundation models~\citep{zhao2023survey,brants2007large, katz1987estimation}. 

% 以下の３段落を１段落にまとめました
After pretraining for general purposes, LLMs are fine-tuned to further better their performance for specific tasks with supervised learning and RLHF -- reinforcement learning from preference labeling feedback from humans~\citep{stiennon2020learning, ouyang2022training}. However, issues arise with RLHF where human feedback can become costly.
% and the feedback can be biased from receiving feedback only from a small group of people. 
To work around this problem, Reinforcement Learning from AI Feedback (RLAIF) was proposed~\citep{bai2022constitutional, lee2023rlaif}, which replaces human feedback with inexpensive feedback from other LLMs.

\begin{figure}
    \centering
    \includegraphics[width=0.9\textwidth]{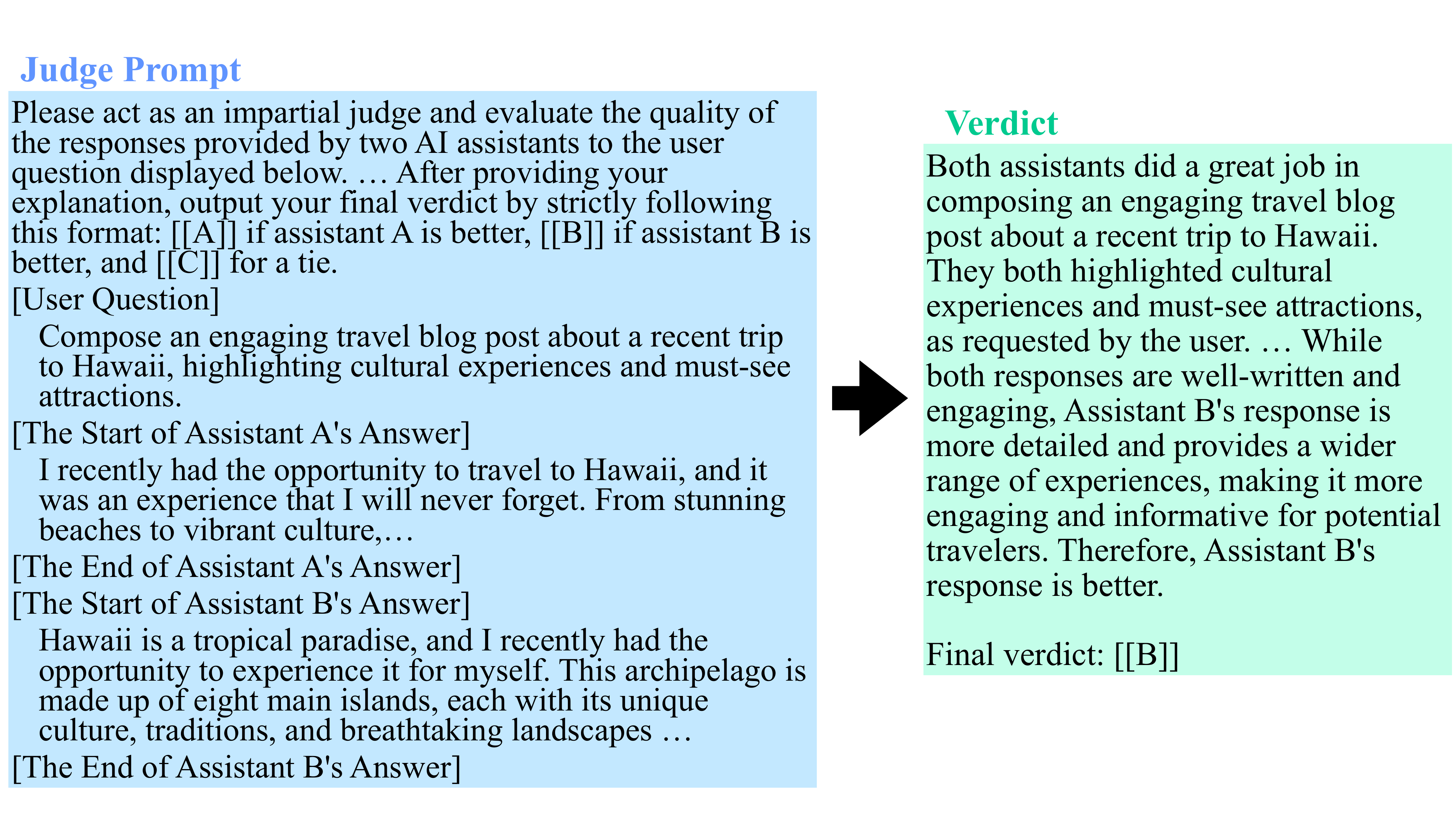}
    \caption{An example of a prompt to an LLM to judge two texts and the verdict. There is no one correct answer, and a comprehensive judgement is required.}
    \label{fig:RLAIF}
\end{figure}

% Tasks without a clear "correct answer" is hard to evaluate, and biases happen (with image of an example task?)
In many cases, the question lack a clear-cut ``correct answer'' and require creativity and imagination. As evident in an example of feedback by an LLM provided in \Cref{fig:RLAIF}, when LLMs are tasked to assess responses to such prompts, the evaluation process can become arbitrary and introduce various biases. One prominent bias is the verbosity bias, which occurs when LLMs are influenced by verbosity, favoring longer and more verbose texts, even if they appear wordy or of lower quality. Without accounting for this bias, LLM agents may learn to generate unnecessarily long texts. This may result in failures in downstream tasks such as lengthy summarizations or chatbots that return verbose responses to simple questions.

% compared to other studies
While previous studies have explored the concept of verbosity bias, they have tended to focus on specific cases. \citet{zheng2023judging} limits their problem setting to questions answered with lists in their experiment on verbosity bias, and \citet{huang2023embrace} conducted experiments on summarization tasks. Moreover, these do not compare the preferences of LLMs to those of humans. We believe that such a comparison is crucial in challenging the conjecture that longer answers are inherently better and that LLMs are actually correct in their preferences.

\paragraph{Our contributions.}
In this paper, we conduct experiments on verbosity bias and saw that 1) LLMs exhibit a preference for longer answers in creative writing tasks, and 2) there is a discrepancy between of LLMs and those of humans in verbosity preference. Additionally, we formulate a quantification for measuring verbosity bias based on accuracy parity. This can be used to compare LLMs on their degree of verbosity bias.

\section{Preliminaries}

% For specific tasks, LLMs are trained with pre-training + fine tuning
After undergoing pretraining for general purposes, LLMs are fine-tuned to further improve their performance in specific tasks. Pretraining is accomplished through self-supervised learning, where the model is trained to predict the next token in a sentence. Once the LLM is able to generate cohesive sentences, we proceed to fine-tune the model to solve specific tasks. One approach to fine-tuning involves supervised learning using expert data. This method relies on examples where experts have solved the task at hand. An example of a conversational LLM trained solely using this approach is Vicuna~\citep{vicuna2023}. Vicuna achieved performance comparable to ChatGPT by utilizing user-shared conversations with ChatGPT as expert data. However, it is worth noting that obtaining expert data is often challenging.
%, and using data from another LLM as expert data can lead to a lack of diversity among LLMs.

% Another is via human evaluation Feedback a.k.a. RLHF like chat-gpt, but that comes with high costs and societal problems
RLHF addresses the challenge of limited training data in supervised learning by leveraging human feedback~\citep{stiennon2020learning,ouyang2022training}. This approach not only mitigates data scarcity but also significantly enhances alignment with human preferences, a critical factor in applications such as question answering. In RLHF, a reward model is trained to closely match human feedback data, which acts as the reward signal in the subsequent RL phase. Prominent LLMs like ChatGPT and Bard adopt a hybrid approach, combining both supervised learning and RLHF techniques to further refine their alignment with human preferences.

% % RLAIF is proposed, which uses evaluation feedback from LLMs instead of humans
% Human feedback comes with its problems. Although human feedback is less expensive than generating expert data from scratch, it is still costly. There is also a risk of an inherent bias from receiving feedback only from a small group of people.
% To work around these problems, Reinforcement Learning from AI Feedback (RLAIF) was proposed \citep{bai2022constitutional, lee2023rlaif}. In RLAIF, human feedback is replaced with much cheaper feedback from other LLMs.

\subsection{RLHF}

\begin{figure}
    \centering
    \includegraphics[width=0.9\textwidth]{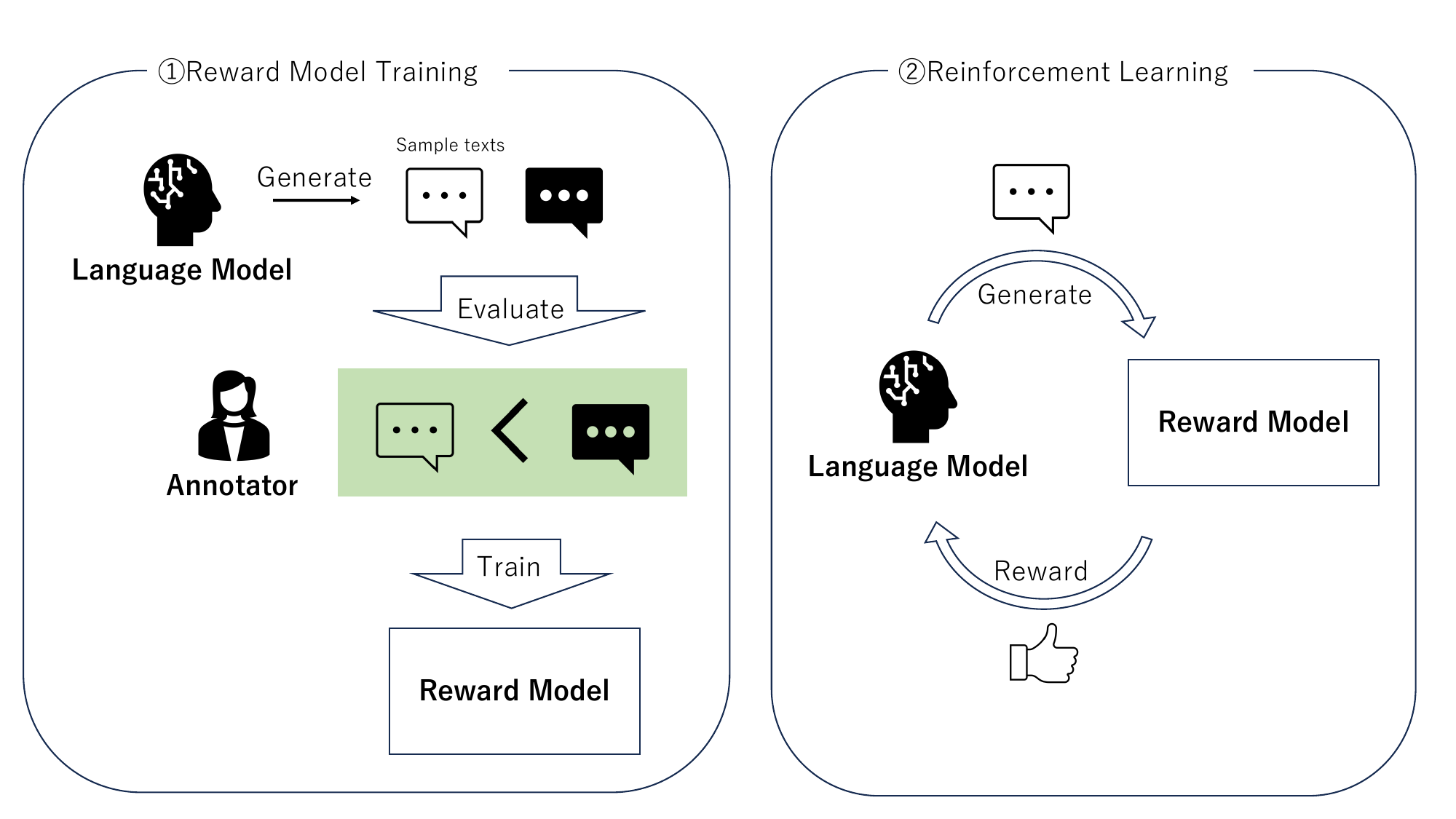}
    \caption{Two phases in RLHF. First the reward model is trained to align with human preference by matching with human feedback. In the second RL phase, the trained reward model provides reward signals to the language model.}
    \label{fig:rlhf}
\end{figure}

% % RLHF is used when fine-tuning LLMs for a specific task. Supervised learning to fine-tune is possible only if expert data exists, and is sometimes combined with RLHF (e.g. ChatGPT).
% When fine-tuning LLMs for a specific task, the natural course of action would be supervised learning from expert data of the task. Either to replace the lack of expert data or for extensive fine-tuning in addition to supervised learning, the LLM is trained with RLHF to align with human preferences.
% RLHFは教師あり学習の〜を解決する手法である...

% First we train the reward model that models human preference. (include equation)
The first step of RLHF is to fit the reward function to align with human feedback. RL directly from human feedback as reward signal is 
% not differentiatable, 
unstable and requires volume. Therefore, a reward model that acts as the reward signal later in the process is trained to be consistant with human preference. Given a dataset $\mathcal{D}$ consisting of the original question, a pair of generated text, and the human preference label on which is chosen and rejected, the reward model is trained by minimizing 
%\eqref{eq:loss_rew_model}.
%
\begin{align}
    \mathcal{L}(\phi) = -\underset{{(x, y_\text{chosen}, y_\text{rejected}) \sim \mathcal{D}}}{\mathbb{E}} \left[\,\log \sigma (r_\phi (y_\text{chosen} \mid x) - r_\phi (y_\text{rejected} \mid x))\, \right],
    \label{eq:loss_rew_model}
\end{align}
where $x$ is the prompt to the LLM, $y_\text{chosen}$ is the preferred text, $y_\text{rejected}$ is the rejected text, and $r_\phi$ parametrised with $\phi$ is the reward model that takes text as input and outputs the rating score.

% Then we train the LLM to maximize reward signal from reward model, like regular RL. (include equation)
In the second step of RLHF, now that we have a reward model to evaluate a generated text without human interaction, regular RL can take place. In this context, the state is the question and the generated text so far, action is the next token to generate, and the reward is $r_\phi(y)$ given after the full text is generated. This is equivalent to a task where a sparse reward is given only at episode termination. The LLM maximizes the signal from the reward model with the following:
\begin{align}
    \max_\theta \mathbb{E} [\,r_\phi (\pi_\theta(x) \mid x)\,],
\end{align}
where $\pi_\theta$ is a policy parameterized by $\theta$. An optional KL divergence term is added to penalize the policy from deviating from the original policy.

\subsection{RLAIF}
% Human feedback is costly/problematic
While RLHF brings down the cost of human labor compared to generating an expert data from scratch, human feedback is still costly. For example, in ~\citet{wang2023large}, it cost around 3 minutes (\$0.75 if \$15.00 per hour) per evaluation. In one occasion, OpenAI worked around this by employing people in Kenya on less than \$2 per hour pay in the process of labeling violent or innappropriate texts. They were under scrutiny for the unideal working conditions.

% RLAIF is proposed, using feedback from another language model instead of humans (include equation)
To combat these promblems, RLAIF was proposed. This method replaces human feedback with feedback from other LLMs. This brings down the cost significantly; in our case, evaluation cost around \$0.05 each, which is 1/15th compared to human feedback in the previously cited paper~\citep{wang2023large}.

\subsection{Biases in Automated LLM Evaluation}

When LLMs evaluate generated texts, various biases are introduced. We provide below a list of biases discussed in various papers~\citep{zheng2023judging, bai2022training, wang2023large}.

\textbf{Position Bias}: Position bias occurs when, in comparing generated texts, LLMs prefer the answer given in certain positions. If we define the ground truth probability of $a$ (the first parameter) preferred over $b$ (the second parameter) to be $P(a, b)$, it should be that $P(y_0, y_1) = 1 - P(y_1, y_0)$, meaning the position should have no affect on the judgement. Position bias is when the comparison by model is $\hat{P} (y_0, y_1) \neq (1 - \hat{P} (y_1, y_0))$. For example, GPT-4 tends to prefer the first option given to it, while ChatGPT prefers the second option~\citep{wang2023large}. To account for this bias, we can simply swap the positions and evaluate the options twice. If the model gives contradicting results between permutations, we count it as a draw.

% This introduces a noise that increases draws. The rate at which this happens can be calculated given the ground truth $P(y_0, y_1)$ and the position bias $\epsilon$ (assuming position bias for each model is only dependant on position and neither $y_0$ nor $y_1$). The rate at which a perfect judge contradicts itself with swapped positions is given below.
% \begin{align}
%     P(y_0, y_1) P(y_1, y_0) + (1 - P(y_0, y_1)) (1 - P(y_1, y_0))
%     = 2 P(y_0, y_1) (1 - P(y_0, y_1))
%     \label{eq:actual_draw}
% \end{align}
% Let $\hat{P} (y_0, y_1) = P(y_0, y_1) + \epsilon$ and $\hat{P} (y_1, y_0) = P(y_1, y_0) + \epsilon$. The rate at which the model 
% \begin{align}
%     \hat{P}(y_0, y_1) \hat{P}(y_1, y_0) + (1 - \hat{P}(y_0, y_1)) (1 - \hat{P}(y_1, y_0)) \nonumber \\
%     % = 2 \hat{P}(y_0, y_1)\hat{P}(y_1, y_0) - \hat{P}(y_0, y_1) - \hat{P}(y_1, y_0) + 1 \nonumber  \\
%     = 2 (P(y_0, y_1) + \epsilon) (1 - P(y_0, y_1) + \epsilon) + 2\epsilon \nonumber \\
%     = 2 P(y_0, y_1) (1 - P(y_0, y_1)) + 2\epsilon^2.
%     \label{eq:model_draw}
% \end{align}
% The draw rate error between \eqref{eq:actual_draw} and \eqref{eq:model_draw} is $2\epsilon^2$, meaning at this rate the evaluations that should have been conclusive is intead judged as draws. Compared to the "aggresive approach" in \citet{wang2023large} which randomizes position and only evaluating the answers once per pair, this approaches decreases the expected error from $\epsilon$ to $2\epsilon^2$ in exchange for evaluation cost (evaluating swapping positions doubles the evaluation cost).

\citet{wang2023large} has proposed several methods to calibrate this bias further: Multiple Evidence Calibration asks the LLM to provide evidence before making judgement, and Human-in-the-Loop Calibration involves human adjustment when deemed neccessary.

\textbf{Self-enhancement Bias}: LLMs tend to prefer answers generated by itself compared to answers generated by other models. This becomes a problem when benchmarking LLMs by evaluating them with LLMs~\citep{zheng2023judging}, but not so much in the context of RLAIF, as the comparisons are always between answers generated by the same model.

\textbf{Verbosity Bias}: Verbosity bias refers to the bias where LLMs prefer longer, more verbose answers even if there are no difference in quality. Training with RLAIF with verbosity bias present can lead to LLMs generating excessively long responses, when in reality a much more concise response would suffice. In tasks such as question answering, a verbose response can be critical to its usefulness, but there aren't enough researches that look into this. For these reasons we take a closer look into this.

There are several proposed methods to mitigate the effect of biases.

\textbf{Chain-of-thought Prompting} is a prompting technique where the LLM is asked to provide the thought process before generating the actual evaluation. This way, at the time when the LLM generates the actual evaluation, it has its chain-of-thought to base its evaluation from. This encourages human alignment and more accurate evaluations, rather than arbitrary evaluations without thought.

\textbf{One-shot/Few-shot Prompting} is another prompting technique which gives one example/several examples of a prompt and its corresponding correct answer when promting the LLM. When generating the response, the LLM can continue the pattern from the examples to better align with the intended response.

\section{Related Works}

% TODO: check https://github.com/MLGroupJLU/LLM-eval-survey

% RLHF:
    % Deep Reinforcement Learning from Human Preferences
    % Reinforcement Learning Atari and Mujoco tasks from human feedback.
    % B-Pref: Benchmarking Preference-Based Reinforcement Learning
    % Benchmark for RLHF algorithms

\subsection{RLAIF Advancements}
There have been several recent advancements in the field of RLAIF. \citet{bai2022constitutional} trained an LLM via RLAIF with limited human feedback. In this work, they claim that helpfulness and harmfulness have a trade-off relationship, and aim to train an LLM that keeps a balance between those two. Their method only requires human feedback in the helpfulness aspect, and harmless behavior is achieved purely from RLAIF. The LLMs trained in the work by \citet{lee2023rlaif} achieved near-human performance in summarization tasks with RLAIF without any human feedback. While not a study on RLAIF itself, \citet{zheng2023judging} evaluate LLMs with other LLMs as a judge and show that GPT-4 has a high human alignment and agrees with humans on over 80\% of evaluations. 
% Error Analysis Prompting Enables Human-Like Translation Evaluation in Large Language Models: A Case Study on ChatGPT
% LLM-EVAL: Unified Multi-Dimensional Automatic Evaluation for Open-Domain Conversations with Large Language Models
% Unified benchmark for LLM evaluation.

\subsection{On Verbosity Bias in Evaluations by LLMs}
\citet{zheng2023judging} also provides lists of biases and methods to overcome them. Alongside their experiment on position bias, they experimented on verbosity bias by attempting a "repetitive list attack" on several LLMs. This attack pertains to "listing" tasks, in which the prompt asks to list several items (e.g. "What are examples of fruits that are round?"). The "repetitive list attack" is done by making the answers verbose by repeating items multiple times, and then asking the LLMs to evaluate these augmented answers. If the LLM evaluates these "repetitive lists" to be better than the original, the attack is considered a success. Their results show GPT-4 is significantly less prone to this attack with below 10\% success rate, while GPT-3.5 and Claude-v1 both suffer over 90\% success rate. Compared to this research, we expand the problem setting to general question-answering tasks.
\citet{huang2023embrace} tackle verbosity bias in summarization tasks. They found that GPT-4 actually prefers short responses in faithfulness and coverage when it comes to summarization, although this is seen strongly only in single-answer grading, and not in comparison grading. This suggests that verbosity bias can be different between different tasks.

Compared to these studies, our problem setting is more general and we compare the verbosity preference between humans and LLMS. The experiments conducted in these papers measure the difference in evaluations when the texts are artificially made verbose while maintaining the same content. The assumption is that elongating the texts would have no effect on a true evaluator, so the difference in evaluation indicates verbosity bias. In our attempt to broaden the problem setting, we make use of human feedback as the oracle instead of making this assumption.
% Large Language Models are not Fair Evaluators (https://arxiv.org/pdf/2305.17926.pdf)
% Attempts to calibrate positional bias with multiple evidence calibration, balanced position calibration, and human-in-the-loop calibration
% Others
% Other works research specific social biases based on gender, etc.

\section{Verbosity Preference of LLMs}
\label{sec:verbosity_preference_llm}

% first we experiment on verbosity bias 
First, we experiment to see how much LLMs actually prefer longer answers. We ask GPT-4 to choose between pairs of responses and examine if it prefers longer responses or not. We did not limit our scope to prompts answered with a specific format (like lists in \citet{zheng2023judging}) in order to observe the LLMs' general tendency to prefer longer answers.

% We generated 100 sample answers to a single prompt ("write a blog about Hawaii...") generated by same model (vicuna), then used GPT-4 to judge these answers. 
We generated 100 sample answers each to 3 prompts, all from the same model (Vicuna-7b-v1.5) generated with the temperature parameter set to 0.7. One of the questions and two examples of the answers are as follows.
\begin{itemize}
    \item Question: \textit{Compose an engaging travel blog post about a recent trip to Hawaii, highlighting cultural experiences and must-see attractions} 
    \begin{enumerate}
        \item I recently had the opportunity to travel to Hawaii, and it was an experience that I will never forget. From stunning beaches to vibrant culture, there was so much to explore and enjoy during my time on the islands...
        \item Hawaii is a tropical paradise, and I recently had the opportunity to experience it for myself. This archipelago is made up of eight main islands, each with its unique culture, traditions, and breathtaking landscapes. During my trip, I had the chance to visit several cultural sites, such as the Polynesian Cultural Center on Oahu...
        \item ...
    \end{enumerate}
\end{itemize}
The prompts are taken from the library introduced by \citet{zheng2023judging}, all from the "creative" category because 1) answers generated to other categories didn't vary in word count enough to see verbosity bias and 2) GPT-4 was not good at judging answers in those categories. We then take answers from these generated samples and insert them into the template shown in \Cref{fig:RLAIF}. With the template complete, we asked GPT-4 to evaluate preferences between pairs of answers with the template. The outcome is either the first option selected, a draw, or the second option selected. In order to account for position bias, GPT-4 evaluated the pair twice with the position swapped the second time. It was considered a draw unless it gave the same result on both permutations.

% \begin{figure}
%     \centering
%     \includegraphics[width=0.6\textwidth]{plots/experiment_1/comparison-all.pdf}
%     \caption{Results with all questions combined. }
%     \label{fig:exp_1_all}
% \end{figure}
\begin{figure}
    \centering
    \begin{subfigure}[]{0.6\textwidth}
    \centering%
    \includegraphics[width=\textwidth]{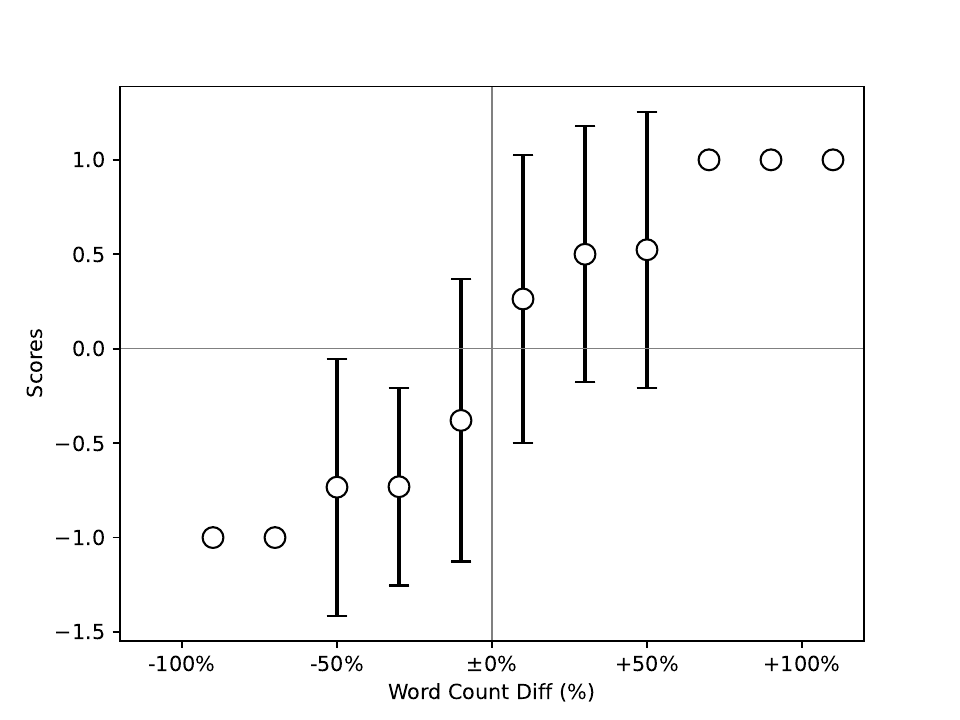}%
    \caption{Results with all questions combined. }%
    \label{fig:exp_1_all}%
    \end{subfigure}%
    \\
    \begin{subfigure}[b]{0.33\textwidth}%
        \centering%
        \includegraphics[width=\textwidth,clip,trim=10 10 20 20]{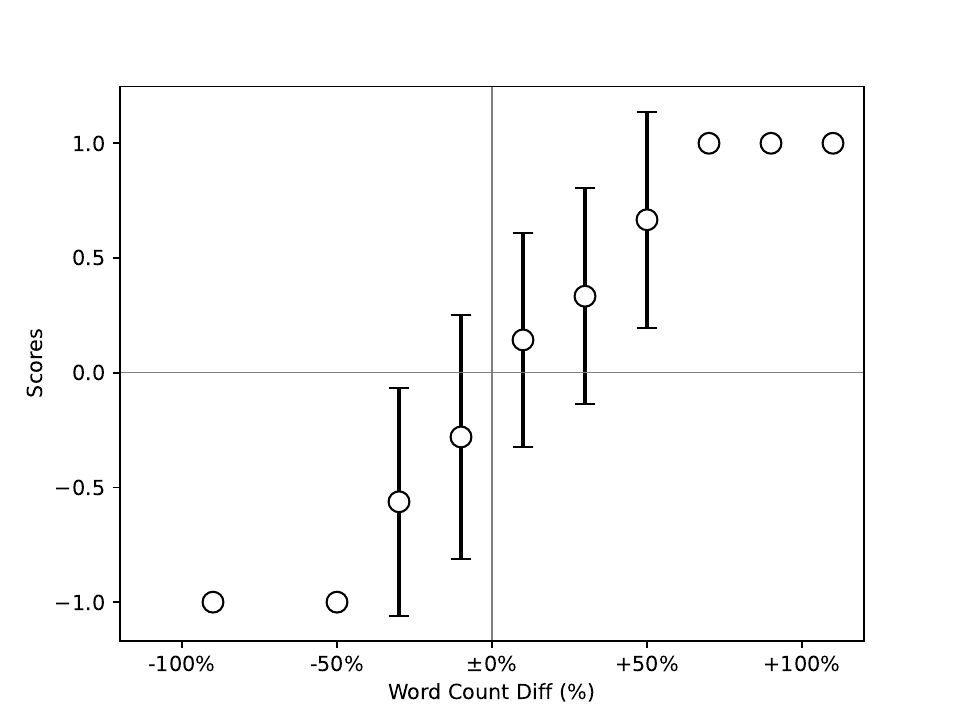}%
        \caption{Question 1}%
       \label{fig:exp_1_indiv1}%
    \end{subfigure}%
    \begin{subfigure}[b]{0.33\textwidth}%
        \centering%
        \includegraphics[width=\textwidth,clip,trim=10 10 20 20]{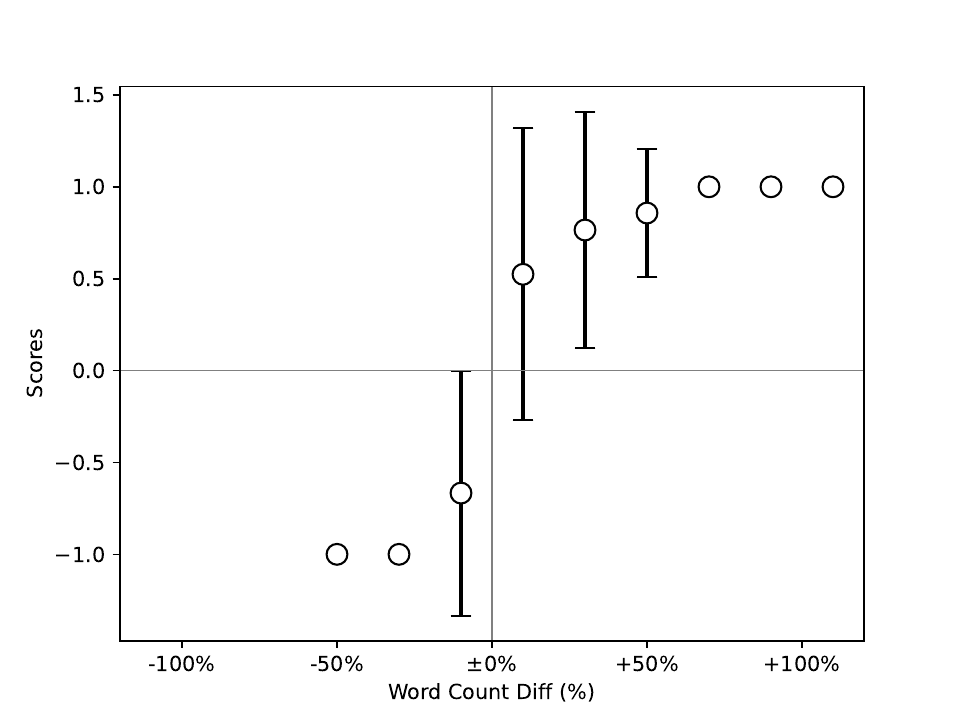}
        \caption{Question 2}%
       \label{fig:exp_1_indiv2}%
    \end{subfigure}%
    \begin{subfigure}[b]{0.33\textwidth}%
        \centering%
        \includegraphics[width=\textwidth,clip,trim=10 10 20 20]{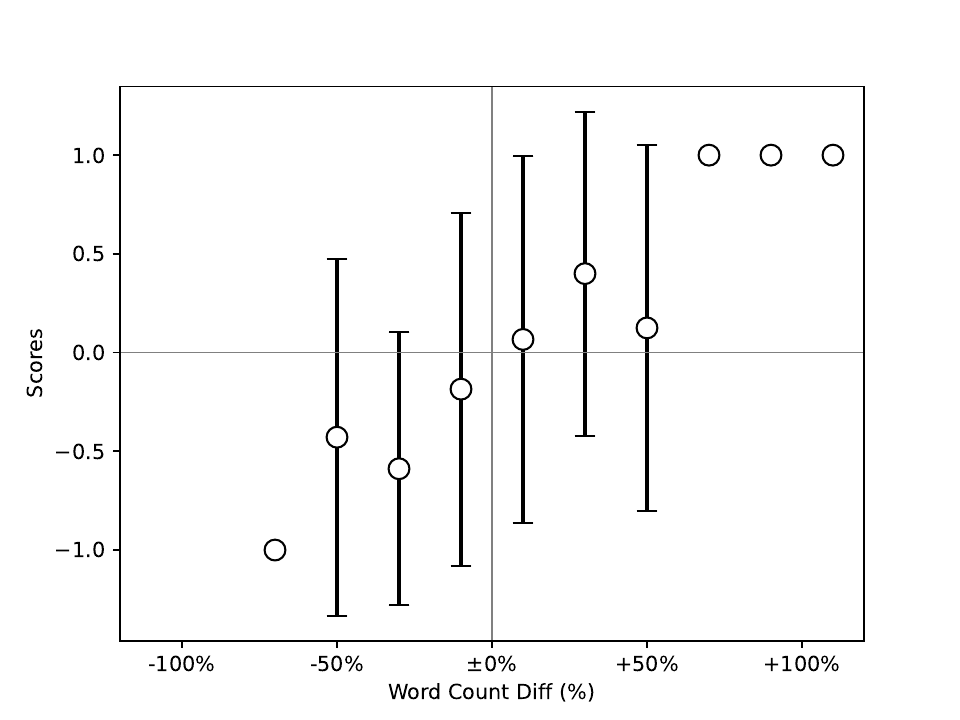}%
        \caption{Question 3}%
       \label{fig:exp_1_indiv3}%
    \end{subfigure}%
    \caption{Example experimental results for three questions (Question 1: Blog about trip to Hawaii, Question 2: Email to professor about paper, Question 3: Blog comparing smartphones). \Cref{fig:exp_1_all} combines results from all three questions and \Cref{fig:exp_1_indiv1,fig:exp_1_indiv2,fig:exp_1_indiv3} show results for each indivisual quesion. X-axis is the percentage of the difference between the first and the second option compared to the length of the second option. Y-axis is the actual score, with 1.0 meaning the first option was selected, 0 meaning it was a draw, and -1.0 meaning the second option was selected. There is an positive correlation between word count difference between the two options and the resulting evaluation. The data points are binned with each ranging 20\%. The circles represent the average in each range and the errorbars show the standard deviation.}
\end{figure}
% \begin{figure}
%     \centering
%     \begin{subfigure}[b]{0.33\textwidth}%
%         \centering%
%         \includegraphics[width=\textwidth,clip,trim=10 10 20 20]{plots/experiment_1/comparison-81.pdf}%
%         \caption{Question 1}%
%     \end{subfigure}%
%     \begin{subfigure}[b]{0.33\textwidth}%
%         \centering%
%         \includegraphics[width=\textwidth,clip,trim=10 10 20 20]{plots/experiment_1/comparison-82.pdf}
%         \caption{Question 2}%
%     \end{subfigure}%
%     \begin{subfigure}[b]{0.33\textwidth}%
%         \centering%
%         \includegraphics[width=\textwidth,clip,trim=10 10 20 20]{plots/experiment_1/comparison-83.pdf}%
%         \caption{Question 3}%
%     \end{subfigure}%
%     \caption{Example experimental results for three questions (Question 1: Blog about trip to Hawaii, Question 2: Email to professor about paper, Question 3: Blog comparing smartphones). X-axis is the percentage of the difference between the first and the second option compared to the length of the second option. Y-axis is the actual score, with 1.0 meaning the first option was selected, 0 meaning it was a draw, and -1.0 meaning the second option was selected. There is an positive correlation between word count difference between the two options and the resulting evaluation. }
%     \label{fig:exp_1_indiv}
% \end{figure}

% experiment 1 results
The results are shown in \Cref{fig:exp_1_all} for the overall result, and \Cref{fig:exp_1_indiv1,fig:exp_1_indiv2,fig:exp_1_indiv3} for results from each prompt. Both in the overall result and the individual results, there is a tendency for GPT-4 to prefer longer answers. When the word count difference is large enough, GPT-4 almost always prefers the longer answer. For question 1 and 2, the preference is smooth and clear, while for question 3, when the word count difference is small, there is high variance in evaluation. As we can see the shape varies between questions, and therefore we can deduce that verbosity does not rely entirely on word count and is different for each question. This makes adjusting for verbosity post-evaluation hard unless we know the verbosity preference shape for the prompt in question.

From this experiment, we can draw the conclusion that GPT-4 generally prefers longer answers among those that are generated by the same LLM with the same prompt. However, this experiment by itself does not indicate that GPT-4 suffers from verbosity bias; it could be that the longer answers generated by vicuna are actually higher in quality and helpfulness. In order to truly measure verbosity bias, we would need the ground truth of each comparison which we do not have. Instead, we next utilize a dataset of human evaluations as the baseline.

\section{Is There A Difference in Verbosity Preference Between LLMs and Humans?}
\label{sec:difference_in_verbosity}

\begin{figure}
    \centering
    \includegraphics[width=0.6\textwidth]{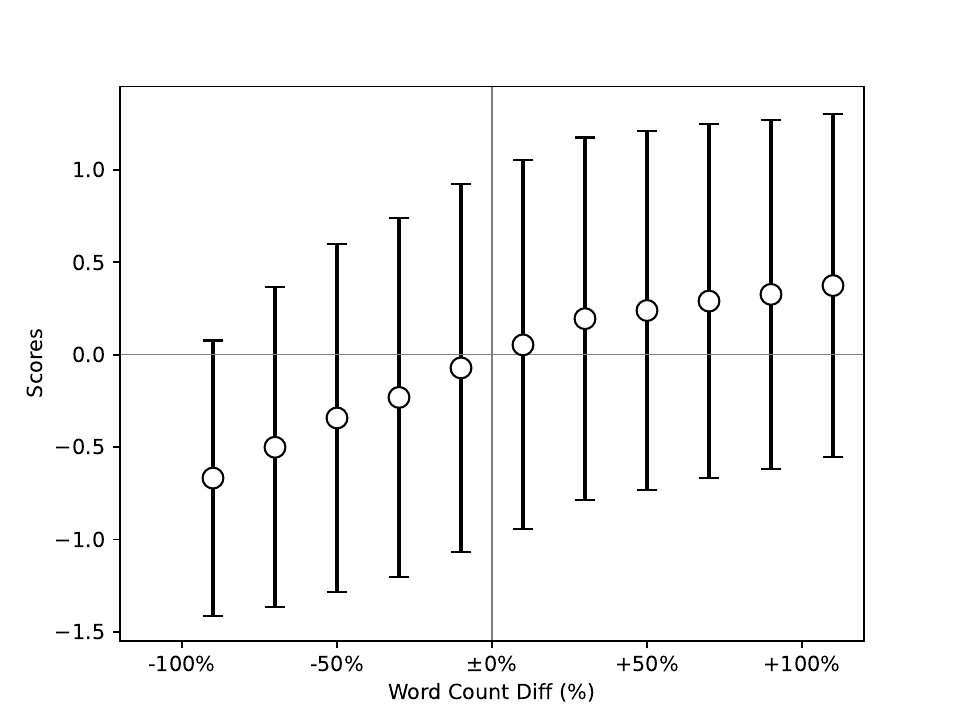}
    \caption{X-axis is the percentage of the difference between the first and the second option compared to the length of the second option. Y-axis is the resulting score by human judgement.}
    \label{fig:human-verbosity}
\end{figure}

% We compare verbosity preference between humans and LMs. Even if we suppose humans, like LMs, have a preference on how long answers should be, it is possible that the preference doesn't completely match. 
Considering that LLMs replace humans as annotators in RLAIF, it is sufficient if LLMs could replicate human feedback and it does not necessarily have to be aligned with the ground truth. As seen in \Cref{fig:human-verbosity} which plots verbosity preference of humans in the HH-RLHF dataset described later, humans seem to prefer longer answers too. Whether or not the longer answers are actually helpful is irrelevant as long as the LLM and the human come to the same conclusion. In light of this, we compare the difference in verbosity preference between LLMs and humans. We can view this as verbosity bias since the aim of LLM judgment in RLAIF is human alignment and not the eradication of biased preference in verbosity.

% We use the hh-rlhf dataset (https://huggingface.co/datasets/Anthropic/hh-rlhf), which contains human feedback data comparing two answers to a prompt. It only has one feedback data per prompt, so we cannot see the verbosity preference like in the experiment in previous chapter.
% we measure human alignment (how often LMs agree with humans in judgement) across various prompts, each with different word counts. 
We use the HH-RLHF dataset \citep{bai2022training} which contains human feedback data comparing pairs of answers to a prompt. It only has one feedback data per prompt, so we cannot plot the verbosity preference of humans like in the experiment in the previous chapter. Instead we can see the dissimilarity between LLMs and humans in verbosity preference in general across various questions. Precisely, this experiment looks into the relationship between the difference in a number of words in the pair of responses and the human alignment of LLMs, meaning how often LLMs give the same judgment as humans.

% (precise experiment details)
We used the same prompt template as the previous experiment but asked GPT-4 to evaluate the whole conversation. Unlike the previous experiment which evaluated answers to a single question, HH-RLHF contains conversations between a human and an assistant. Therefore we asked GPT-4 to evaluate the pair of whole conversations and answer which assistant was more helpful.
% this might be why humans prefered longer answers too -- te longer the conversation, thhe more the assistant is engaging?

\begin{figure}
    \centering
    \begin{subfigure}[b]{0.49\textwidth}
        \centering
        \includegraphics[width=\textwidth]{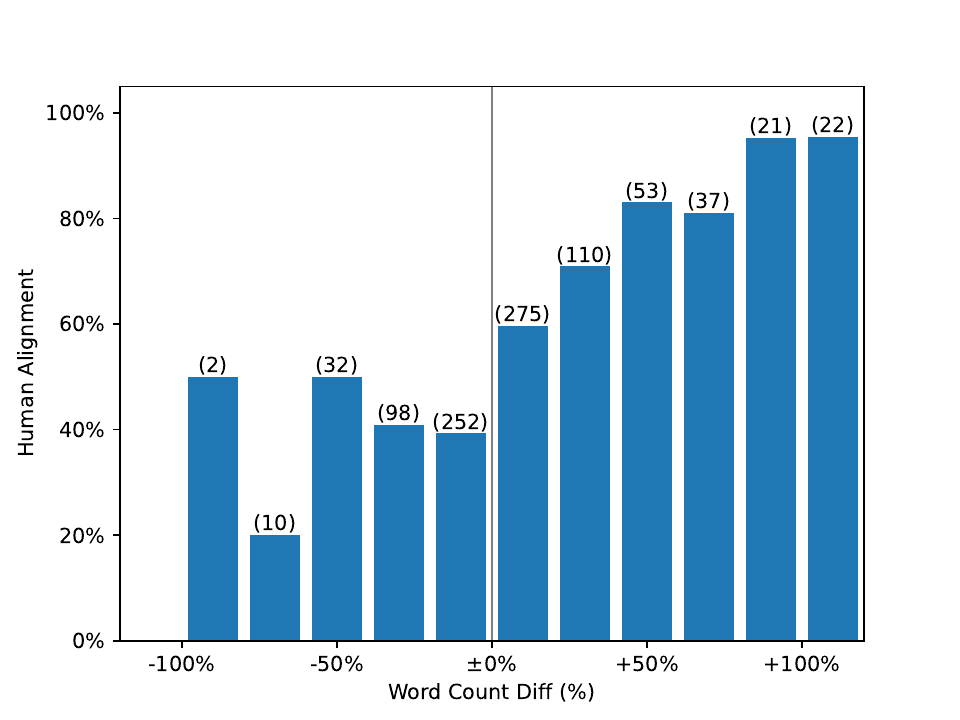}
        \caption{GPT-4}
        \label{fig:exp_2_gpt-4}
    \end{subfigure}
    \begin{subfigure}[b]{0.49\textwidth}
        \centering
        \includegraphics[width=\textwidth]{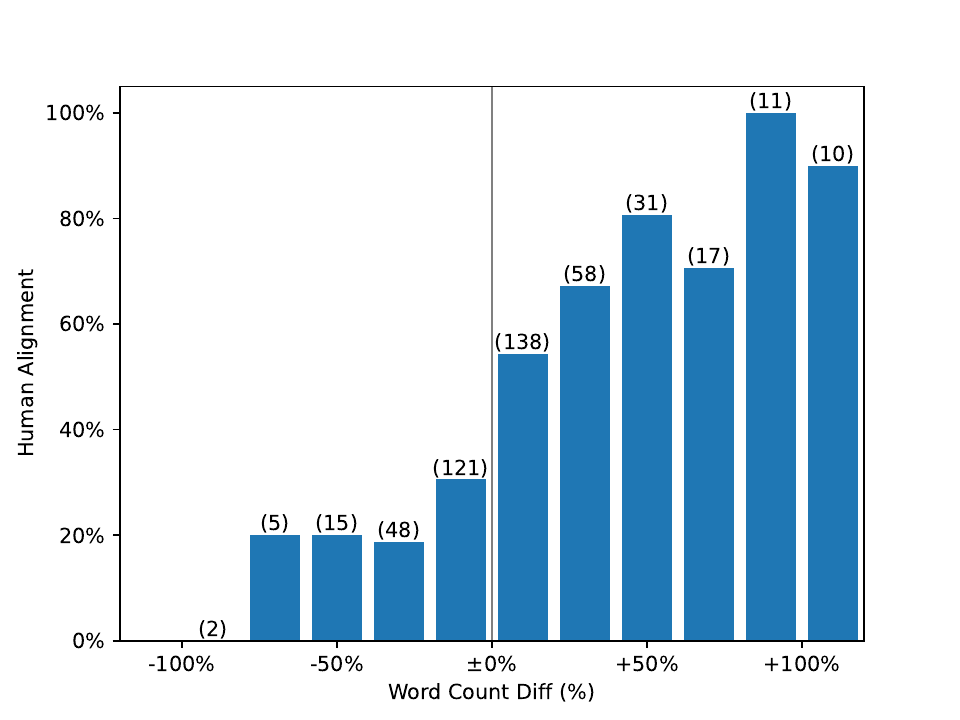}
        \caption{GPT-3.5}
        \label{fig:exp_2_gpt-3.5}
    \end{subfigure}
    \caption{X-axis is the percentage of the difference between the chosen and the rejected option, compared to the length of the rejected option. Y-axis is the human alignment measured by the rate of LLM's decision agreeing with humans. The numbers in brackets indicate the sample size in each bracket.}
    \label{fig:enter-label}
\end{figure}

% In the cases where human feedback preferred the answer with more words, human alignment was high, meaning GPT-4 preferred that answer as well. However, when human feedback chose the answer with less words, human alignment was low, because GPT-4 couldn't choose the same answer.
In cases where human feedback preferred the longer answer, human alignment was high for the LLMs, meaning the LLMs preferred the longer answers as well. However, when human feedback chose the answer with fewer words, human alignment was low, because the LLMs still chose the longer answers regardless of the helpfulness of the shorter answer.  

% One possible explanation is that LMs learned to mimic human behavior heuristically by choosing longer answers. In this dataset, human feedback did tend to favor longer responses. 
One possible explanation for this is that LLMs learned to mimic human behavior heuristically by choosing longer answers -- in this dataset, human feedback did tend to favor longer responses as seen in \Cref{fig:human-verbosity}, and it is possible the dataset used to train GPT-3.5/GPT-4 had the same tendency. Nevertheless, a closer look into the cause is up for debate.

\section{Formulation of Verbosity Bias}
\label{sec:formulation}

% In the second experiment we saw the tendency of LLMs to be less accurate when the human prefers shorter answers. We formulate verbosity bias
In the second experiment, we observed the tendency of LLMs to have low human alignment for cases where human feedback preferred shorter answers. In this section, we formulate verbosity bias to allow for quantative comparison between models.

In our problem setting, we define the given pair of text inputs as $y_0$ and $y_1$, the LLM outputs decision as $Y' \in \{0, 1\}$, and the more helpful option labeled by humans as $Y \in \{0, 1\}$. We define the sensitive attribute $S \in \{0, 1\}$ which equals 0 when $y_0$ has more words than $y_1$, and 1 when $y_1$ has more words than $y_0$. 

With these definitions, equal opportunity~\citep{hardt2016equality} with respect to sensitive attribute $S$ is satisfied if 
\begin{align}
    P(Y'=0 | S = 0, Y = 0) = P(Y'=0 | S = 1, Y = 0).
\end{align}

This only accounts for cases where human feedback prefers $y_0$. Although this can be attained by sorting the inputs beforehand, the equation can be generalized with accuracy parity instead of equal opportunity. Accuracy parity is satisfied if the accuracy of prediction is equal among both demographics:
\begin{align}
    P(Y' = Y | S = Y) = P(Y' = Y | S = 1-Y).
\end{align}

% \begin{align}
%     P(Y' = Y | S = Y) > P(Y' = Y | S = 1-Y)
% \end{align}

The deviance from accuracy parity can be calculated with the following equation:
\begin{align}
    |P(Y' = Y | S = Y) - P(Y' = Y | S = 1-Y)|.
\label{eq:verbosity_bias_abs}
\end{align}

Even though this is how the deviance is calculated in general, we thought it important that the directional information of the bias isn't lost. With the formulation below \eqref{eq:verbosity_bias}, a positive value indicates that the LLM prefers verbose answers, and a negative value indicates it prefers shorter answers. This distinction is crucial as some tasks may have a negative bias, for example in summarization tasks as shown in~\citet{huang2023embrace}. We also opted for the difference in \textit{inaccuracy} between demographics 
\begin{align}
    P(Y' = 1 - Y | S = 1-Y) - P(Y' = 1 - Y | S = Y)
\label{eq:verbosity_bias}
\end{align}
because verbosity bias refers to the inacuracy influenced by verbosity.

\Cref{table:verbosity_values} shows the verbosity bias values of GPT-3.5 and GPT-4 calculated with data from \Cref{sec:difference_in_verbosity}. From these numbers, we can conclude that GPT-4 has improved in verbosity bias. Compared to~\citet{wang2023large}, which had a limited problem setting and gave the impression that GPT-4 is significantly less prone to verbosity bias, we see that the verbosity bias still exists for GPT-4. A further experiment on other LLMs for comparison is required.

\begin{table}[t]
\caption{Verbosity bias values calculated with \eqref{eq:verbosity_bias} for GPT-4 and GPT-3.5 with data from experiment.}
\begin{center}
\begin{tabular}{@{}lll@{}}
\toprule
Model          & GPT-4 & GPT-3.5 \\ \midrule
Verbosity Bias & 0.328 & 0.428   \\ \bottomrule
\end{tabular}
\end{center}
\label{table:verbosity_values}
\end{table}

% exact values: 
% GPT-4: 0.3283577088673334
% GPT-3.5: 0.4275402836374727

% % From these results, GPT-4 has improved in verbosity bias. Compared to [3], result from broader problem setting
% This result suggests GPT-4 is suffers less that GPT-3.5 in verbosity bias, while not completely invulnerable. Though the result in \citet{} showed GPT-4 to be significantly less affected by "repetitive list attack", it is up to debate whether GPT-4 is as robust in the general setting.

\section{Discussion}

% in order to mitigate this, it needs to be fixed during the training?

% importance sampling / RLHF is similar to offline RL, but does not account for how LLM output has different distribution between before and during RLHF phase

% In this context, the sensitive attribute is verbosity. Different from normal sensitive attributes because verbosity SHOULD have effect on prediction.
\subsection{Other Metrics of Equality}
In the context of our study, we treat the verbosity of the response pair as the sensitive attribute in our formulation of verbosity bias in \Cref{sec:formulation}. What verbosity differs from sensitive attributes generally discussed in other cases of biases is the fact that verbosity \textbf{should} actually be taken into consideration when evaluating the responses, whereas attributes like gender or race \textbf{shouldn't} be a factor in the outcome in other cases. This is why employing other metrics of equality like demographic parity doesn't make sense here, and therefore we base the measurement of verbosity bias on equal opportunity and accuracy parity.

% 研究自体のLimitation
\subsection{Limitations of Our Experiments}
In the experiment in \Cref{sec:verbosity_preference_llm}, we generate the sample responses from the same questions (before concatenation of results from all three questions). However, in our experiment in \Cref{sec:difference_in_verbosity}, we mix together results from various questions. This has led us to only attain the result across many kinds of questions, not the result on any specific question like in the experiment in \Cref{sec:verbosity_preference_llm}. It is debatable which of these results is preferable.

% our formulation cannot account for bias between groups within each side. for example, a verbosity bias with a peak around +-0% would have small verbbosity bibas value. graphs shhould be used alongside these values
\subsection{Limitation of Our Metric of Verbosity Bias}
Our formulation of verbosity bias only accounts for bias between two groups divided by whether $y_0$ is longer than $y_1$. What it cannot detect is the bias within each of these groups; it is agnostic to the bias between cases where $y_0$ is barely longer than $y_1$ and cases where $y_0$ is significantly longer than $y_0$. Hence, if there were to be an instance where the model has high human alignment when there is a large difference in length between the pair of responses -- the plot would have a concave shape symmetrical around the vertical line down the middle -- our metric would suggest that the model has close to zero verbosity bias. To avoid such a situation, showing the human alignment plot alongside the metric is recommended.

\section{Conclusion}
% In previous papers, ...
% In our experiment, we found...
In this paper, we conducted experiments on the verbosity bias seen in LLMs' judgment by LLMs. In previous works, the problem settings were limited and did not compare the verbosity preference to humans. With our experiments, we saw that 1) LLMs tend to favor longer answers for creative writing tasks, and 2) alignment with humans varies on verbosity with lower human alignment in cases where humans preferred shorter answers. We then formulated verbosity bias based on accuracy parity that can be used to quantitatively compare verbosity biases among models.

%\section*{References}

% \newpage

\small

% \bibliographystyle{apalike}
% \bibliography{neurips_2023}

\appendix

\end{document}